# Non-linear stiffness behavior of planar serial robotic manipulators

Wanda Zhao[1], Alexandr Klimchik[2], Anatol Pashkevich[1,3], Damien Chablat[1,4]

[1] *Laboratoire des Sciences du Numérique de Nantes (LS2N), UMR CNRS 6004,* Nantes, France

[2] *Innopolis University, Universitetskaya St, 1, Innopolis, Tatarstan 420500, Russia*

[3] *IMT Atlantique Nantes, 4 rue Alfred-Kastler, Nantes 44307, France*

[4] *Centre National de la Recherche Scientifique (CNRS), France*

Wanda.Zhao@ls2n.fr, A.Klimchik@innopolis.ru, Anatol.Pashkevich@imt-atlantique.fr, Damien.Chablat@cnrs.fr

## ABSTRACT

The paper focuses on the stiffness analysis of multi-link serial planar manipulators, which may demonstrate nonlinear stiffness behavior under the compressive loading. Two important cases are considered, where the manipulator has either a straight or non-straight initial configuration. It was proved that in the first case the loading may cause the buckling if it exceeds some critical value, and the manipulator suddenly changes its straight shape and stiffness properties. For computing this critical force, a general eigenvalue-based technique was proposed that can be applied to any multi-link serial manipulator. For the second case dealing with non-straight initial configurations, a universal energy-based technique was applied that allowed to detect quasi-buckling phenomenon when it is observed very fast but not instant change of the manipulator shape and its stiffness coefficient under the loading. These results are illustrated by numerous examples of non-linear stiffness behavior of three- and four-link manipulators that are subjected to compressive force.

# Nomenclature

| | |
|---|---|
| $F$ | the external force |
| $q$ | the mechanism configuration angle |
| $α$ | an angle defining the mechanism initial shape |
| $k$ | the stiffness coefficient of the rotational spring |
| $L$ | the length of the links |
| $F_C$ | the critical force |
| $V$ | the potential energy |
| $U$ | the strain energy |
| $\Delta$ | the end-effector deflection |
| $(x_0, y_0)$ | the end-point coordinates |
| $\mathbf{F}=[F_x\ F_y]$ | the external force, |
| $\mathbf{M}$ | the internal torques |
| $\mathbf{J}$ | the Jacobian matrix |
| $q_1, q_2$ | the mechanism configuration angles (for 3-link mechanism) |
| $E(.)$ | the strain energy of this mechanism |
| $U^+, U^-, Z^+, Z^-$ | the manipulator configuration shapes |
| $(\delta_x, \delta_y)$ | the end-effector displacement |
| $\mathbf{S}_0$, $\mathbf{S}_1$ | the geometric parameter matrices |
| $\mathbf{K}_q = diag(k_1,...,k_n)$ | the joint stiffness matrix. |
| $\mathbf{A}$, $\mathbf{B}$ | $(n+1)\times(n+1)$ square matrices of parameters |
| $F_x^0$ | the desired critical force |
| $\{\mathbf{v}_1,...,\mathbf{v}_{n+1}\}$ | the eigenvectors |
| $v_{ij}$ | the components of the eigenvector $\mathbf{v}_i = \begin{bmatrix} v_{i,1},...,v_{i,n+1} \end{bmatrix}^\mathrm{T}$ |
| $\lambda_i$ | the eigenvalues |
| $\mu$ | some small scaling factor |

# 1 Introduction

Mechanical stiffness is one of the most important indicators in the performance of industrial robotic systems [1-3]. For instance, for machining applications where the primary target is the precise manipulation of a technological tool, the stiffness defines the positioning errors due to the external loading caused by the manipulator link weights or interaction with the workpiece. Similarly, in industrial pick-and-place applications which are intended for simple but fast manipulations, the stiffness defines an admissible range of velocity/acceleration while approaching the target point, in order to avoid undesirable positioning errors due to inertia forces [4]. The stiffness is also very important in medical applications with large robotic manipulators for patient positioning, where elastic deformations of mechanical components under the task load are not negligible [5]. It is obvious that in all of these cases, the manipulator links/joints should be rigid enough to ensure the high stiffness required by the relevant application. On the other hand, there are currently a number of fields where the required manipulator stiffness is rather low to ensure safe interaction with a human operator. Such compliant manipulators are already widely used both in industrial automation and medical equipment due to their flexibility, modularized construction, low weight and ability to excuse sophisticated motions. However, for some of such manipulators, their stiffness behavior under the loading is quite different from conventional mechanisms that are usually treated as multi-dimensional springs with constant parameters described by 6×6 stiffness matrices [6]. For instance, under loading, the compliant manipulators may suddenly change their shape or partially lose stiffness. Such phenomenon (also known as mechanical buckling) has been already been reported by several researchers, but general methods of nonlinear stiffness analysis for robotic manipulators have not been created yet. This paper concentrates on some of such issues and proposes a new technique allowing to detect buckling phenomenon in serial compliant robotic manipulators subjected to the axial loading.

Currently, the stiffness analysis of robotic manipulators is mainly based on three popular methods, which are the Finite Element Analysis (FEA), the Virtual Joint Method (VJM), and the Matrix Structural Analysis (MSA) [7-10]. The most accurate of them is the **FEA** [11-13], which allows modeling links and joints with their true dimension and shape. However, this technique is usually applied at the final design stage because of the high computational expenses required for high order matrix inversion [14, 15]. In contrast, the **VJM** method is treated as the simplest one, it is based on the extension of the traditional rigid model by adding the virtual joints (localized springs), which describe the elastic deformations of the links, joints and actuators [16-19]. This technique is widely used for serial and strictly parallel robots, but it can be hardly applied to manipulators with more complex topology. The **MSA** is considered as a compromise technique, which incorporates the main ideas of the FEA, but operates with rather large elements such as flexible links connected by the actuated and passive joints in the overall manipulator structure [20-23]. This obviously leads to the reduction of the computational expenses that are quite acceptable for robotics, but it requires some non-trivial actions for including passive and elastic joints in the related mathematical model. Despite their high popularity, the above-mentioned three methods provide the user with linear stiffness models allowing them to compute only small manipulator deformations under the external loading. Their main instrument is the 6×6 stiffness matrices that relate the external forces/torques with the linear/angular deflections of the manipulator end-effector. These matrices are usually assumed to be constant and there is a rather limited number of works that investigate their dependence on the loading amplitude [23-25]. On the other hand, in structural mechanics, the nonlinear stiffness behavior had been already carefully studied for some typical elements such as columns, beams, thin plates, etc. The most famous example is the Euler-Bernoulli theory of compressed column, which keeps its initial straight shape under the small loading, but suddenly bends or buckle if this load is larger than some critical value [26-29]. But for robotics, the buckling phenomenon was rarely studied before because usually engineers prefer to exclude buckling while designing a robot. There are only several related works presenting some case studies dealing with three-link serial manipulators or serial mechanisms based on tensegrity structures [30, 31]. Nevertheless, for compliant manipulators, this phenomenon is important and must be anticipated on the desired stage, which requires **the** development of special nonlinear stiffness analysis techniques.

The main objective of this paper is to develop a general technique for serial planar manipulators allowing us to compute the critical force causing the buckling phenomenon. In contrast to the previous work, which concentrated on some case studies, this new technique should be applicable to any multi-link manipulator that is subjected to compressive loading in its straight configuration. In addition, some other nonlinear phenomena will be investigated, such as quasi-buckling for initial non-straight configuration, when it is observed very fast change (but not instant) of the manipulator shape and its stiffness coefficient under the loading.

To address these issues, the remainder of this paper is organized as follows. Section 2 gives some theoretical background on buckling in continuous and discrete systems. Section 3 concentrates on the manipulator's stiffness analysis for non-straight initial configuration and presents a universal energy-based technique allowing to detection of quasi-buckling phenomena in the considered mechanical structures. Section 4 contains the main theoretical contribution, a general eigenvalue-based method for the critical force computing for straight initial configuration of the multi-link serial manipulators under axial loading. Finally, Section 5 summarizes the main results and contributions.

## 2  Buckling in discrete mechanical systems and robotics

In structural engineering, one of the main nonlinear stiffness phenomena is known as buckling, which refers to the loss of stability of a component, when it suddenly changes its shape under the loading. Such behavior is mainly known for the beams and columns, which spontaneously bend from straight to curved form under a compressive load [32-34]. However, this event can be also observed in some discrete mechanical systems including robotic manipulators [35-38]. The buckling phenomenon was detected in a number of mechanisms composed of rigid/flexible links and springs, which are connected by passive or elastic joints. Let us consider the first two-link serial mechanisms connected by a rotational spring shown in **Fig. 1**.

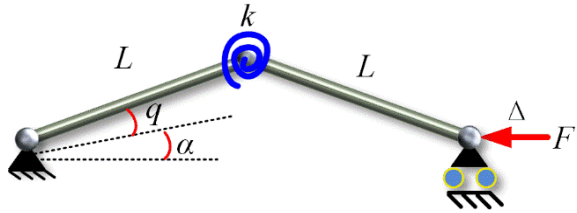

**Fig. 1** A two-link mechanism where the buckling phenomenon is observed

For this mechanism, the static equilibrium equation is $FL\sin(\alpha+q)=2kq$, which allows us to get the desired force-deflection relation $F(\Delta)$ in the following parametric form.

$$F(q)=2kq/L\sin(\alpha+q); \quad \Delta(q)=2L\left(\cos\alpha-\cos(\alpha+q)\right), \qquad (1)$$

where $F$ is the external force, $q$ is the mechanism configuration angle, $\alpha$ is an angle defining the mechanism initial shape, $k$ is the stiffness coefficient of the rotational spring and $L$ is the length of the links. An example of this force-deflection curve for the case $\alpha=0$ (i.e. for the straight initial configuration) is presented in **Fig. 2a**, which shows that at first the function $F(\Delta)$ goes from the zero to a non-zero value without changing the deflection $\Delta$, until achieving the critical force causes the buckling. Then the force $F$ is increasing monotonically as the deflection $\Delta$ is increasing. The exact value of the critical force for this mechanism in straight configuration ($\alpha=0$) can be found by computing the limit of $F(q)$ for $q\to 0$:

$$F_C=\lim_{q\to 0}F(q)=2k/L; \quad \text{for } \alpha=0 \qquad (2)$$

The above buckling phenomenon can be also analyzed using the energy method, by considering maximums and minimums of the potential energy curves corresponding to the stable and unstable equilibriums respectively [39, 40]. The potential energy $V$ of the considered mechanism can be expressed as $V(q)=U(q)-F\cdot\Delta(q)$, where $U$ is the strain energy stored in the spring, $F$ is the applied conservative load, and deflection $\Delta$ is the distance moved by $F$ in its direction. Here the strain energy $U=k\cdot(2q)^2/2$, so the potential energy $V$ can be expressed as

$$V(q)=2kq^2-2FL\left(\cos\alpha-\cos(\alpha+q)\right). \qquad (3)$$

For the case $\alpha=0$ the energy curves on the plane $(V, \Delta)$ are presented in **Fig. 2b**, which shows that for $F\le F_C$ the energy minimum is achieved when Δ=0 and the equilibrium is stable and corresponds to the straight shape of the mechanism ($q$=0). In contrast, for $F > F_C$ the energy minimum is achieved when Δ≠0, which corresponds to the non-straight mechanism shape in the stable equilibrium state. It is clear that this shape depends on the external force $F$.

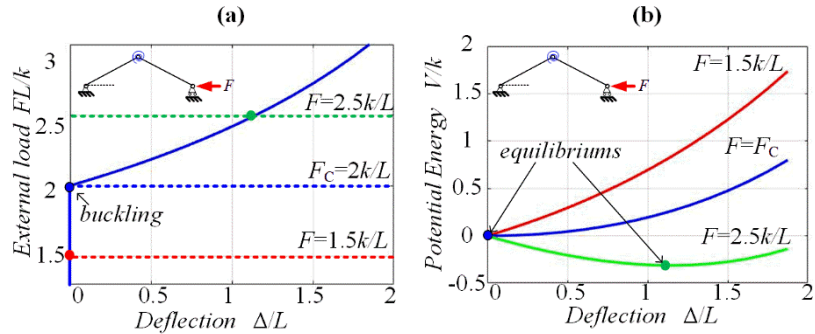

**Fig. 2** Straight initial configuration: the force-deflection and energy curves for two-link mechanism (case $\alpha$=0).

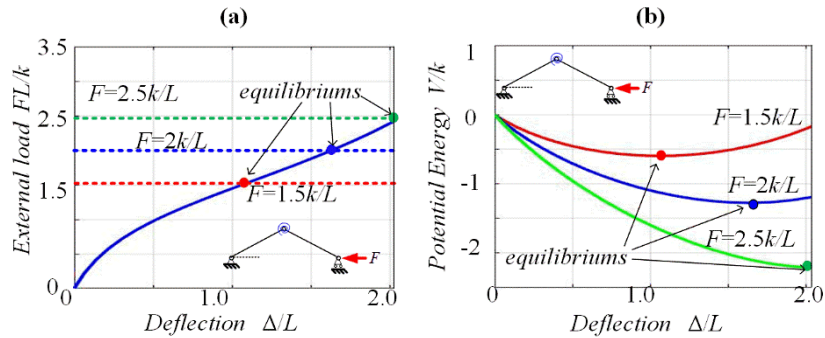

**Fig. 3** Non-straight initial configuration: the force-deflection and energy curves for two-link mechanism (case $\alpha$=π/6).

For the case of $\alpha$≠0, i.e. the non-straight initial configuration, the above equations produce continuously monotonic force-deflection curves passing through the zero point $F(\Delta)$=(0, 0) as shown in **Fig. 3a**. So, the mechanism behavior is similar to a simple spring that always resists external loading. Corresponding energy curves are shown in **Fig. 3b**, all of them have a single minimum defining stable equilibrium with different Δ≠0 depending on the external loading $F$. Hence, the considered two-link mechanism with the rotational spring may demonstrate the so-called "bifurcation" buckling similar to the Euler's column.

Further, let us also consider a three-link serial mechanism shown in **Fig. 4**. The left endpoint is connected to the base by a passive joint, there are two linear springs located on the second and the third joints, and the right endpoint is also equipped with a passive joint and is moved horizontally by the external force $F$. Here $L$ denotes the lengths of the three similar rigid links, $k$ denotes the stiffness coefficient of the two rotational linear springs. It is also assumed that the orientation angle of the first link is denoted as $\varphi$, and the initial values of the second and the third angles are $\alpha_1$ and $\alpha_2$ respectively. Thus, under the loading $F$, the deflection Δ depends on the orientation angles $\alpha_1+q_1$ and $\alpha_2+q_2$, and the end-point coordinates $(x_0, y_0)$ for the initial configuration can be expressed as

$$x_0 = L\cos\varphi + L\cos(\varphi+\alpha_1) + L\cos(\varphi+\alpha_1+\alpha_2)$$
$$y_0 = L\sin\varphi + L\sin(\varphi+\alpha_1) + L\sin(\varphi+\alpha_1+\alpha_2)$$
(4)

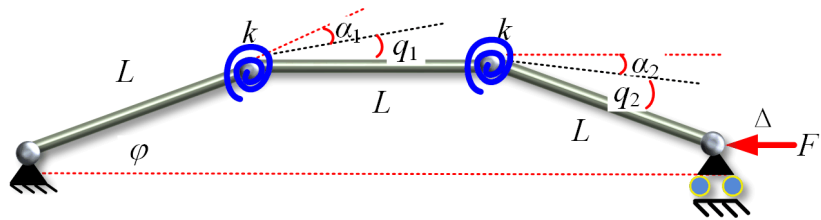

**Fig. 4** Example of three-link mechanism with the buckling phenomenon.

Then under the loading, the end-point moves to a new position

$$x = L\cos\varphi + L\cos(\varphi + \alpha_1 + q_1) + L\cos(\varphi + \alpha_1 + q_1 + \alpha_2 + q_2)$$
$$y = L\sin\varphi + L\sin(\varphi + \alpha_1 + q_1) + L\sin(\varphi + \alpha_1 + q_1 + \alpha_2 + q_2) \quad (5)$$

that can be also expressed via the deflection $\Delta$ as $x = x_0 - \Delta$; $y = 0$. It should be noted that here, because of the geometric constraints $y_0 = y = 0$. To find the angles $q_1$, $q_2$ corresponding to the external force $F$, it is necessary to consider the static equilibrium equation that generally is written as follows

$$\mathbf{J}^T \mathbf{F} + \mathbf{M} = \mathbf{0}, \quad (6)$$

where $\mathbf{F} = [F_x \; F_y]$ is the external force, $\mathbf{M}$ is the internal torques in the joints that can be computed via the angles $q_1$, $q_2$ as $\mathbf{M} = [0 \; kq_1 \; kq_2]^T$ and $\mathbf{J}$ is the Jacobian matrix of size 2×3 that is expressed as

$$\mathbf{J} = L \cdot \begin{bmatrix} -S_0 - S_{01} - S_{012} & -S_{01} - S_{012} & -S_{012} \\ C_0 + C_{01} + C_{012} & C_{01} + C_{012} & C_{012} \end{bmatrix}. \quad (7)$$

Here for further convenience, it is denoted $S_0 = \sin\varphi$, $C_0 = \cos\varphi$, $S_{01} = \sin(\varphi + \alpha_1 + q_1)$, $C_{01} = \cos(\varphi + \alpha_1 + q_1)$, and $S_{02} = \sin(\varphi + \alpha_1 + q_1 + \alpha_2 + q_2)$, $C_{02} = \cos(\varphi + \alpha_1 + q_1 + \alpha_2 + q_2)$. Taking into account that the static equilibrium corresponds to the minimum of the potential energy, for the known equilibrium configuration described by the angles $(\varphi, q_1, q_2)$, the external force can be obtained from the equation (6) by using the Moore-Penrose pseudoinverse

$$\mathbf{F}(\mathbf{q}) = -\left[\mathbf{J}^T(\mathbf{q})\right]^\dagger \cdot \mathbf{M}(\mathbf{q}), \quad (8)$$

where $(\mathbf{J}^T)^\dagger = \mathbf{J}(\mathbf{J}^T\mathbf{J})^{-1}$, and $\mathbf{q} = (\varphi, q_1, q_2)$. It is clear that for such equilibrium configuration the corresponding end-effector deflection $\Delta(\mathbf{q})$ can be straightforwardly computed from the direct geometric model.

It should be noted that the above equations operate with three variables $(\varphi, q_1, q_2)$, but the dimension of this problem can be easily reduced using the analytical solution of the geometric equations (5) for the end-effector location $(x, y) = (x_0 - \Delta, 0)$. The latter allows us to replace the initial configuration space $(\varphi, q_1, q_2)$ by a reduced space $(\Delta, \varphi)$, which is more convenient for the stiffness analysis. This reduction can be easily executed by applying the inverse kinematics of a two-link serial manipulator that yields the following expressions for the angles $q_1$, $q_2$.

$$q_1 = \operatorname{atan2}(y_0 - L\sin\varphi, x_0 - \Delta - L\cos\varphi) - \operatorname{atan2}(LS_2, L + LC_2) - \varphi - \alpha_1; \quad q_2 = \operatorname{atan2}(S_2, C_2) - \alpha_2, \quad (9)$$

where $C_2 = \left((x_0 - \Delta - L\cos\varphi)^2 + (y_0 - L\sin\varphi)^2 - 2L^2\right)/2L^2$, $S_2 = \pm\sqrt{1 - C_2^2}$. Hence, using the above expressions, a single redundant variable $\varphi$ corresponding to the equilibrium configuration can be found from the given $\Delta$ using condition of the min/max of the spring strain energy, i.e. $\varphi = \arg\min_\varphi E(\Delta, \varphi)$ for a stable equilibrium and $\varphi = \arg\max_\varphi E(\Delta, \varphi)$ for an unstable one, where the strain energy of this mechanism can be expressed as

$$E(\Delta, \varphi) = kq_1(\Delta, \varphi)^2/2 + kq_2(\Delta, \varphi)^2/2. \quad (10)$$

It should be mentioned that here, assuming that $\Delta$ is given, minimization of the full potential energy $V(\Delta, \varphi) = E(\Delta, \varphi) - F_x \cdot \Delta$ is not necessary for computing the redundant variable $\varphi$.

Some simulation results based on the above expressions are presented in **Fig. 5**. They include the mechanism force-deflection curves for different initial configurations defined by parameters $\alpha_1$ and $\alpha_2$. For the initial straight or quasi-straight configuration (see **Fig. 5**) the mechanism stiffness behavior is similar to the compressed Euler's column. In particular, there are also here four equilibriums (U$^+$, U$^-$) and (Z$^+$, Z$^-$) corresponding to the half-sine-shape and sine-shape of the compressed column (see **Fig. 6**). However, only (U$^+$, U$^-$) are stable and either U$^+$ or U$^-$ equilibriums can be observed in practice depending on small initial perturbations of $(\alpha_1, \alpha_2)$. For example, in **Fig. 5b** for relatively small $(\alpha_1, \alpha_2) = (\varepsilon, \varepsilon)$ the initial U$^+$ configuration is the only stable one that can be observed. Hence, in practice, only stable configurations can be observed, which are also referred to as feasible ones while unstable are referred to the unfeasible.

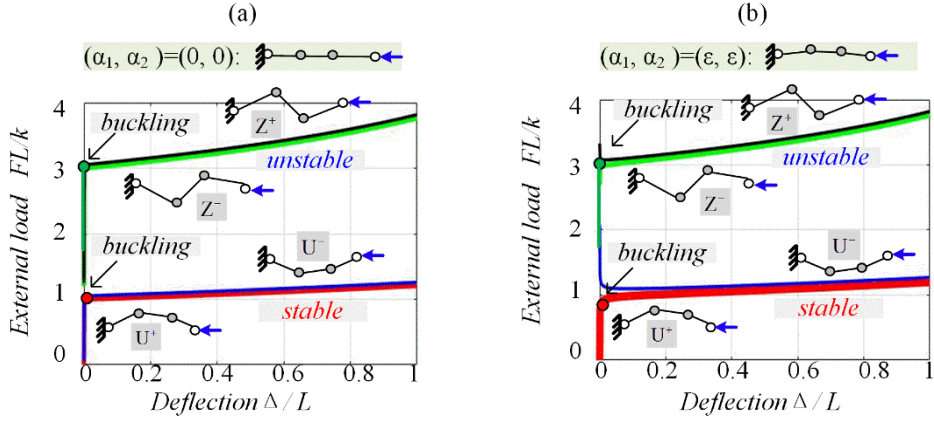

**Fig. 5** Three-link manipulator: force-deflection curves for the initial "straight" and "quasi-straight" configurations and four possible equilibriums with stable shapes ($U^+$, $U^-$) and unstable shapes ($Z^+$, $Z^-$).

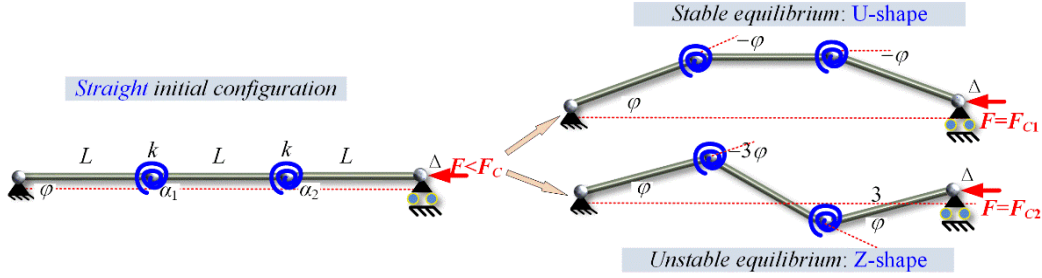

**Fig. 6** Evolution of the mechanism shape after the buckling: two possible equilibriums with U-shape and Z-shape.

Therefore, the buckling phenomenon can be observed both in continuous and discrete mechanical systems such as robotic manipulators. In both cases the mechanism behavior is roughly similar: for the initial straight configuration, the mechanism keeps its shape if the compressive loading is smaller than some critical value. Then it suddenly changes the shape and further the shape is changing continuously while the loading is increasing. However, in the general case, the problem of critical force computing for such discrete mechanisms is still challenging. To our knowledge, there is no universal technique allowing **us** to compute the critical force for any given structure. In literature there are only several works dealing with two- and three-link manipulators, where the critical force was obtained using the energy method [40, 41]. Another challenging issue is related to the initial non-straight configuration, where the manipulator may continuously change its shape under the loading but, after reaching some critical value, its stiffness in a certain direction may be suddenly lost [42]. These problems are in the focus of the following section that concentrates on the serial manipulator under the loading for both singular and non-singular initial configurations.

## 3 Manipulator stiffness in the neighbourhood of the non-straight configuration

As follows from the previous section, for the initial-straight (i.e. singular) configuration, the manipulator has essentially non-linear stiffness behavior under the compressive loading. However, for practice the so-called non-straight initial configuration is more typical, for which the kinematic Jacobian is non-singular. Let us evaluate in detail the manipulator stiffness for this case assuming that the external loading is compressive as in the Euler column. It is intuitively clear that here the manipulator shape under the loading will be changing continuously. But its resistance to the compression described by the stiffness coefficient may essentially depend on the loading amplitude [23, 24]. Below, this issue is investigated for the case of a three-link serial manipulator first, and then this methodology is applied to a general $n$-link case.

### 3.1 Stiffness properties of a three-link manipulator

To estimate properties of the three-link manipulator for the non-straight configuration, let us consider different non-zero values of ($\alpha_1$, $\alpha_2$) describing the initial manipulator shape and compute set of possible equilibrium configurations ($\varphi$, $q_1$, $q_2$) as well as deflections $\Delta$ corresponding to different compressive forces $F$. To simplify computing, it is possible to slightly modify this problem by considering the deflection $\Delta$ as an independent variable and compute corresponding force $F$ and configuration variables ($\varphi$, $q_1$, $q_2$) satisfying the static equilibrium equation (6) and the geometric model (5). This leads to a numerical solution of four nonlinear equations with respect to four variables $\Delta$ and ($\varphi$, $q_1$, $q_2$) that corresponds to the external force $(F_x, F_y) = (F, 0)$ and the endpoint location $x = x_0 - \Delta$, $y = 0$. However, similar to the previous section, it is possible to apply the energy method and find the equilibrium configurations in a straightforward way, by minimizing the manipulator potential energy (10) taking into account geometric constraints (5).

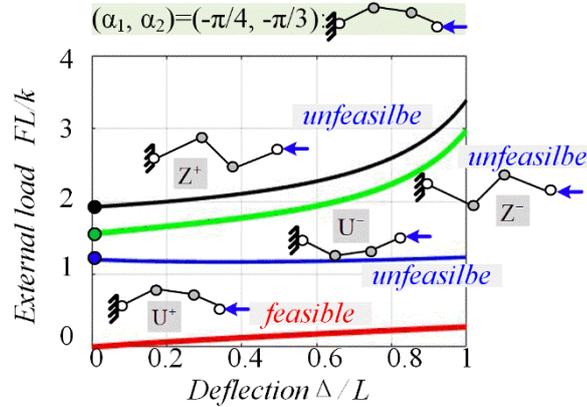

**Fig. 7** Force-deflection curves computed for different shapes and the same initial "U-configuration"
corresponding to four possible equilibriums with the feasible shape $U^+$ and unfeasible shapes ($U^-$, $Z^+$, $Z^-$).

Such an approach yields a one-dimension optimization problem $E(\Delta, \varphi) \to \min$ with respect to the joint angle $\varphi$ assuming that the deflection $\Delta$ is given. It is clear that the latter assumption allows applying the inverse kinematics (9) to compute the joint variables ($q_1$, $q_2$) corresponding to the current value $\varphi$. Using the obtained equilibrium configurations (which are obviously not unique because of $\pm\sqrt{*}$ in the inverse kinematic expressions) it is easy to compute the corresponding compressive force from (8). It is worth mentioning that here in addition to two stable equilibriums corresponding to the energy minimum, there are also two unstable equilibriums corresponding to the maximum of the energy. But in practice, we are only interested in a single equilibrium configuration corresponding to the global minimum of the potential energy.

Computation results corresponding to the initial configuration angles ($\alpha_1$, $\alpha_2$) = ($-\pi/4$, $-\pi/3$) are presented in **Fig 7**. Corresponding manipulator shape with negative ($\alpha_1$, $\alpha_2$) will be further referred to as the $U^+$-*configuration*. For this case, there were obtained four different force-deflection curves corresponding to either the potential energy minimums or maximums. It is worth mentioning that these curves correspond to different manipulator shapes, which are shown in **Fig 7** and denoted as ($U^+$, $U^-$) and ($Z^+$, $Z^-$). It is clear that in practice only a single equilibrium with $U^+$ shape is observed that corresponds to the minimum of the compressive force $F$. Hence, for the considered $U^+$ initial configuration the manipulator shape is changing continuously, there is no buckling phenomenon happened (i.e. no sudden jump from one shape to another).

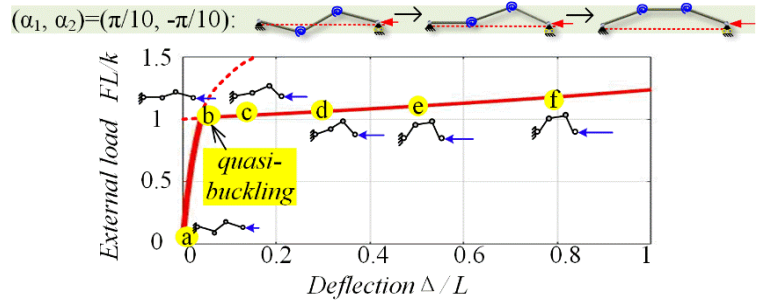

**Fig. 8** Force-deflection curves for the initial "Z-configuration"
and evolution of the manipulator shape under the loading.

In contrast, for the initial $Z^-$-*configuration* with ($\alpha_1$, $\alpha_2$) = ($\pi/10$, $-\pi/10$), the simulation produces essentially different results. Here, the manipulator stiffness behavior has some particularities compared to the above case with the initial $U^+$-*configuration*. In particular, the force-deflection curve corresponding to the potential energy minimum is not smooth (see **Fig. 8**), and the buckling phenomenon happens, when the manipulator stiffness suddenly reduces and its shape suddenly jumps from $Z^-$ to $U^+$. It is worth mentioning that similar to the above case with the initial $U^+$- configuration, here there are also four different equilibriums (two stable and two unstable ones), but only a single curve corresponding to the energy minimum is shown in the figure.

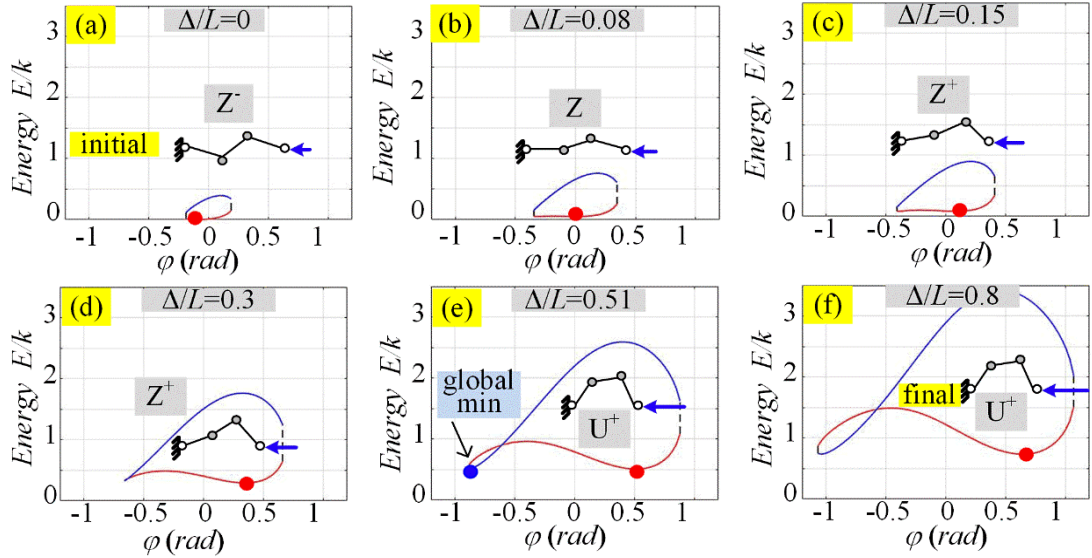

**Fig. 9** Evolution of energy-$\varphi$ curves for stable configuration for the initial "Z-configuration and changing of the manipulator shape under the loading (from $Z^-$ to $U^+$ configuration).

The above presented nonlinear manipulator behavior for the initial $Z^-$-configuration can be also illustrated using the energy curves on the $(E, \varphi)$ plane. Corresponding figures obtained for different deflections $\Delta$ are shown in **Fig 9**. As follows from them, at the beginning the mechanism keeps the initial $Z^-$-configuration with negative $\varphi<0$ (with the angle signs "-+-") and changes its shape continuously (see **Fig. 9a**). Then, after the force $F$ reaches some critical value, the angle $\varphi$ becomes equal to zero $\varphi=0$ (see **Fig. 9b**), and the mechanism stiffness essentially reduces. Further, for the higher loading, this angle becomes positive $\varphi>0$, the mechanism maintains the stable equilibriums $Z^+$-configuration (with the angle signs "+-+" and corresponding to the minimum potential energy) and the deflection changing under the external force is monotonic (see **Fig. 9c, d**). It is worth mentioning that for some configurations the observed manipulator shape corresponds to the local (instead of the global) minimum of the potential energy (as in **Fig. 9e**), but the mechanism is not able to jump to this new configuration, instead, it keeps maintaining the $U^+$-configuration of the local minimum of the potential energy (see **Fig. 9f**). To analyze the stability of equilibrium configuration it is possible to use either the straightforward energy-based technique or the matrix criteria proposed in [43]. In fact, for most practical applications only stable equilibrium configurations are feasible. However, for some geometrical constraints (see [44, 45] for examples) unstable configurations with negative stiffness might be also feasible.

Summarizing all the above case studies dealing with the three-link manipulator, we can conclude that some mechanical structures can demonstrate some new type of non-linear stiffness behavior with a sudden change of the mechanism shape and its stiffness coefficient. In literature, such behavior is well known for initial straight configurations that under the compressive loading can suddenly jump to a curved one, similar to the half-sine shape of the compressed Euler's column. However, in this section, the buckling phenomenon was also detected for some other initial configurations (non-straight ones) that were not studied in Euler's column theory. In particular, it was proved that under the compressive loading the three-link serial mechanism can suddenly change its configuration from the so-called "Z-shape" to "U-shape" and essentially reduce the stiffness coefficients. In the following section, a similar phenomenon will be detected for multilink serial structure based on the tensegrity mechanisms.

## 3.2 Stiffness properties of multi-link manipulator

To demonstrate the existence of nonlinear stiffness behavior in more complicated mechanical structures let us consider a serial multi-link manipulator connected by the elastic joints (**Fig. 10**), where the link lengths $(L_1,...,L_n)$ and the joint stiffness coefficients $(k_1,...,k_n)$. For this case study, we will also use the above-presented energy method, which is also quite efficient here allowing us to detect very interesting phenomena in the manipulator stiffness behavior.

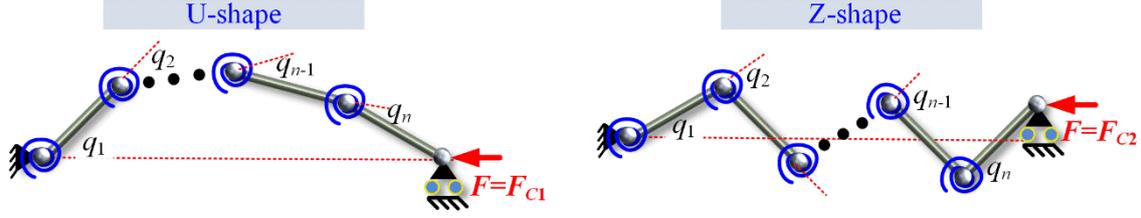

**Fig. 10** The multi-link manipulator in the "non-straight" initial configurations.

**Table 1** Two different initial configurations of the four-link manipulator for the same end-point $(x_0, y_0) = (3.8L, 0)$.

|  | Initial shape | Initial configuration angles (rad) | | | |
|---|---|---|---|---|---|
|  |  | $q1$ | $q2$ | $q3$ | $q4$ |
| Case #1 | U-shape: | −0.3179 | +0.0558 | +0.3804 | +0.3524 |
| Case #2 | Z-shape: | −0.2417 | +0.6821 | −0.7958 | +0.5170 |

Let us assume that the initial configuration of the considered multi-link manipulator is a non-straight one ($q_i^0 \neq 0$, $i = 1, 2, ..., n$) and the initial end-point location is $(x_0, y_0)$, which can be also expressed via the initial longitude displacement $\Delta_x$,

$$x_0 = \sum_{j=1}^{n}\left(L_j \cdot \cos(\sum_{i=1}^{j} q_i^0)\right) = \sum_{j=1}^{n} L_j - \Delta_x \qquad y_0 = 0 \qquad (11)$$

It is clear that if $n \geq 3$ this manipulator is cinematically redundant if the target location is defined by its Cartesian coordinates $(x, y)$ only. Moreover, for given $(x_0, y_0)$ the configuration angles $q_i^0$ also cannot be defined uniquely. For this reason, in the further analysis, we will consider two typical initial manipulator shapes (U-shape and Z-shape), which correspond to the same $(x_0, y_0)$. Examples of such initial shapes for the four-link manipulator are presented in **Table 1**, where it is assumed that $L_i = L$, $\forall i$.

First, let us obtain the force-deflection relations $F_x(\delta_x)$ and $F_y(\delta_x)$ corresponding to the end-effector displacement with $\delta_x = \text{var}$, $\delta_y = 0$, i.e. from the initial location $(x_0, y_0)$ to the current one $(x, y)$. In this case, the geometric constraints derived from the direct kinematics can be presented in the form of two equations

$$x = \sum_{j=1}^{n}\left(L_j \cdot \cos(\sum_{i=1}^{j} q_i)\right) = \sum_{j=1}^{n} L_j - \Delta_x - \delta_x \qquad y = 0, \qquad (12)$$

where $\delta_x$ is the end-effector deflection caused by the external forces $(F_x, F_y)$ and $\Delta_x$ is the initial displacement before the loading. To find loaded equilibriums corresponding to the given $\delta_x$, let us apply the energy method used in the previous section. This allows us to reduce the number of independent variables from $n$ to $n$-2. The latter is achieved by applying the two-link manipulator inverse kinematic expressions similar to (9) and computing the angles $(q_{n-1}, q_n)$ from the remaining ones $(q_1, q_2, ... q_{n-2})$. Further, it is possible to apply a straightforward numerical search to find minimums of the energy function $E(q_1, q_2, ... q_{n-2})$ corresponding to locally stable equilibriums. Then, the obtained configuration variables $(q_1, q_2, ... q_n)$ are substituted to the expression (8), which yields the desired external forces $(F_x, F_y)$ for the given deflection $\delta_x = \text{var}$, $\delta_y = 0$.

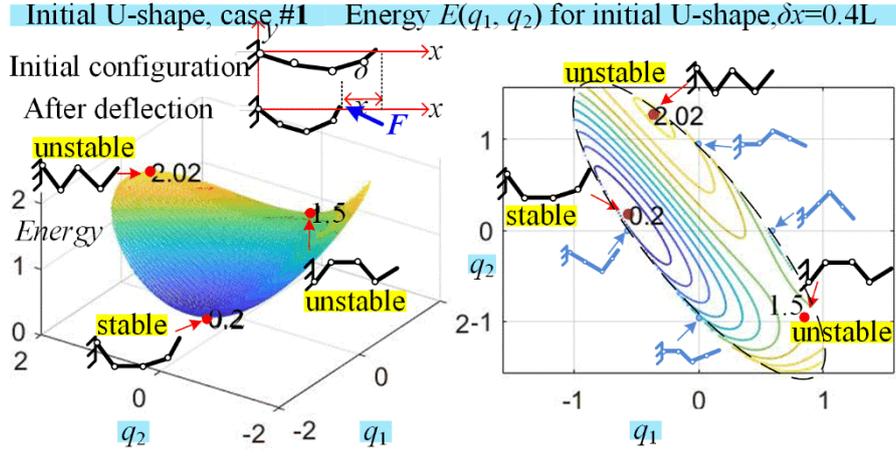

**Fig. 11** The energy function $E(q_1, q_2)$ and equilibrium configurations of four-link manipulator for the initial U-shape (end-effector deflection $\delta x=0.4L$, $\delta y=0$; $q_4>0$).

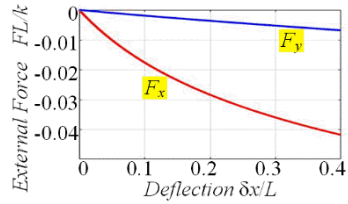

**Fig. 12** Force-deflection curves $F_x(\delta x)$, $F_y(\delta x)$ for initial U-shape (end-effector deflection $\delta y=0$; $q_4>0$).

Examples of the possible equilibrium configurations for loaded four-link manipulators that were obtained using the energy method are presented in **Fig. 11**. Here it is assumed that $L_1 = L_2 = ... = L_n$, $k_1 = k_2 = ... = k_n$ and the manipulator has the initial U-shape with $\Delta_x = 0.2L$ ( see

**Table 1** ) and the deflection caused by the loading is $\delta_x = 0.4L$. For this case, the energy function $E(q_1, q_2)$ has a single maximum, a single minimum and a single saddle point. Corresponding equilibrium configurations are shown in the figure where the stable one has a U-shape (as the initial one). A similar study for other deflections $\delta_x$ shows that the energy function $E(q_1, q_2)$ evolution with respect to $\delta_x$ is continuous, which proves that the force-deflection curves are also continuous as shown in **Fig. 12**. Here the change of the manipulator shape is smooth, the resistance of the manipulator is gradually increasing while the deflection $\delta_x$ becomes bigger. However, the stiffness coefficient in the *x*-direction is continuously decreasing.

Another example presented in **Fig. 13** demonstrates quite different stiffness behavior under the loading. Here the manipulator has an initial Z-shape with $\Delta_x = 0.2L$ (see

**Table 1**) and the deflection caused by the loading is $\delta_x = 0.4L$. For this case, the energy function $E(q_1, q_2)$ has a single maximum and a single minimum point. It is worth mentioning that here the stable point corresponds to the Z-shape (similar to the initial configuration). However, the energy surfaces $E(q_1, q_2)$ are quite different in this case, their evolution with respect to $\delta_x$ has some singularities. In particular, the force-deflection curve $F_x(\delta x)$ shown in **Fig. 14** is not very smooth, it contains a singular area where the stiffness coefficient becomes very low. For this case study, there are two intervals of the manipulator deformation. In the beginning, when $\delta x$ is relatively small the manipulator maintains its Z-shape and the resistance against the external force is increasing fast and monotonically. Further, when the deflection $\delta x$ is bigger than some critical value, the quasi-buckling phenomenon is occurring, and the manipulator resistance against the external force is increasing very slowly. Correspondingly, the stiffness coefficient $dF_x/dx$ becomes very small.

In the above-presented examples, it was assumed that the initial configuration is changed by means of the actuation of corresponding joints. Further, the actuator coordinates remain the same, but the joint angles are changed under the influence of the external force (because of the joint elasticity only).

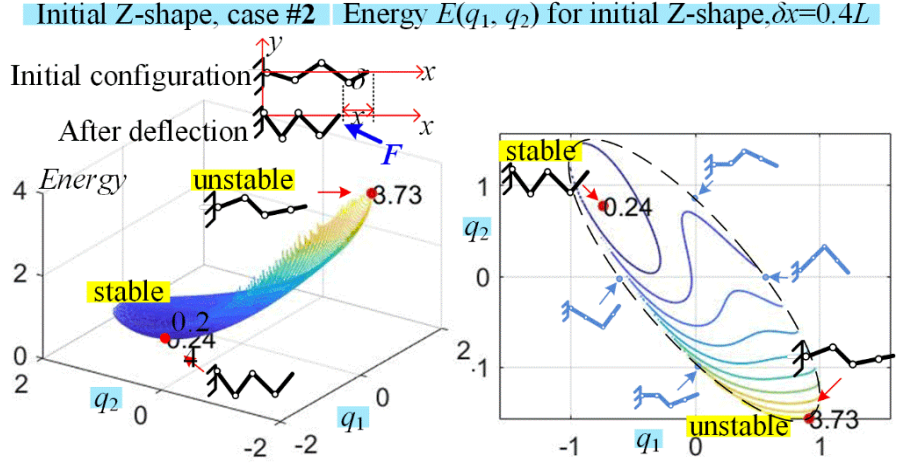

**Fig. 13** The energy function $E(q_1, q_2)$ and manipulator equilibriums of initial Z-shape configuration (end-effector deflection $\delta x/L=0.4$, $\delta y=0$; $q_4>0$).

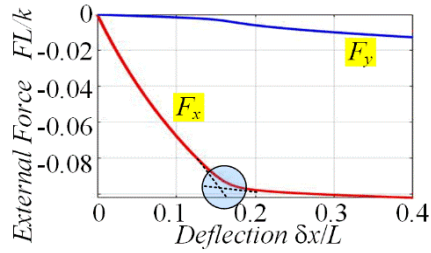

**Fig. 14** Force-deflection curves $F_x(\delta x)$, $F_y(\delta x)$ for initial Z-shape (end-effector deflection $\delta y=0$; $q_4>0$).

It is worth mentioning that similar results can be also obtained by straightforwardly solving the static equilibrium equations (8) with respect to $n$ unknown variables $(q_1,...,q_n)$, i.e. without imposing any kinematic constraints on the end-effector location. In this approach, it is assumed that the external loading $(F_x, F_y)$ is given while the manipulator endpoint deflections $(\delta x, \delta y)$ are computed from the configuration angles $(q_1,...,q_n)$ satisfying the equilibrium equations.

Hence, for this case study dealing with the $n$-link serial manipulator, it was also discovered nonlinear stiffness behavior for non-straight initial configuration, when it is observed very fast change (but not instant) of the manipulator shape and its stiffness coefficient under the loading. It is worth mentioning that usually, such behavior is typical for mechanical structures with the initial straight shape that can suddenly jump to a curved one under the compressive loading. It should be mentioned that in the presented simulation studies only a single quasi-buckling point was discovered. But intuitively it is quite possible that several quasi-buckling points may exist for some particular initial configurations. So it is an open question for further research.

## 4 Manipulator stiffness in the neighborhood of the straight configuration

Now let us consider in detail the manipulator stiffness behavior under the loading assuming that the initial configuration is straight. Intuitively, this behavior should be similar to the compressed Euler column presented at the beginning of this paper, i.e. the manipulator should keep its straight shape until the external loading is larger than a certain critical value. However, computing of this critical force is not a trivial issue and is addressed below.

### 4.1 Critical force for three-link manipulator

Let us present first a specific technique for a relatively simple case, which deals with a three-link manipulator with two elastic joints considered in the previous section (see **Fig 4**). It is assumed that here its initial shape is straight but under the compressive loading, the manipulator changes its shape to a non-straight one. It is clear that after the buckling the manipulator should achieve either "U" or "Z" shape with some symmetrical properties that can be taken into account by defining the configuration angles as $(\varphi, -\varphi, -\varphi)$ or $(\varphi, -3\varphi, 3\varphi)$ correspondingly.

To compute the desired critical force it is necessary to assume that after buckling the above-mentioned angle $\varphi$ is very small, i.e. $\varphi \to 0$. It is clear that either for the "U" or "Z" cases the considered mechanical system must satisfy the usual static equilibrium equation $\mathbf{J}^T\mathbf{F}+\mathbf{M}=\mathbf{0}$, which relates to the external force $\mathbf{F}=(F_x, F_y)^T$, the internal joint torques $\mathbf{M}=(0, kq_1, kq_2)^T$, and the manipulator Jacobian $\mathbf{J}$. It is worth mentioning that here the initial joint angles are $\alpha_1 = \alpha_2 = 0$, and the geometric constraints (5) can be excluded from the consideration because it is taken into account implicitly, by the assumption that the configuration angles are either $(\varphi, -\varphi, -\varphi)$ or $(\varphi, -3\varphi, 3\varphi)$. So, the desired value of critical loading can be computed directly from the static equilibrium equation by computing the limit of the force $\mathbf{F}$ for $\varphi \to 0$.

In the case of the U-shape, the kinematic Jacobian can be expressed as

$$\mathbf{J} = L \cdot \begin{bmatrix} 0 & \sin\varphi & \sin\varphi \\ 1+2\cos\varphi & 1+\cos\varphi & \cos\varphi \end{bmatrix} \tag{13}$$

giving the following equations for the static equilibrium

$$\begin{aligned} L(1+2\cos\varphi)\cdot F_y &= 0 \\ L\sin\varphi\cdot F_x + L(1+\cos\varphi)\cdot F_y - k\cdot\varphi &= 0 \\ L\sin\varphi\cdot F_x + L\cos\varphi\cdot F_y - k\cdot\varphi &= 0 \end{aligned} \tag{14}$$

The solution of the system is

$$F_x = k\varphi/L\sin\varphi; \qquad F_y = 0 \tag{15}$$

and describes the stable force-deflection curve in **Fig. 5a.** So, computing the relevant limit that yields the following critical force for the U-shape case

$$F_x^U = \lim_{\varphi\to 0}\left(\frac{k\varphi}{L\sin\varphi}\right) = \frac{k}{L} \tag{16}$$

Similarly, in the second case (Z-shape), the kinematic Jacobian can be rewritten as

$$\mathbf{J} = L \cdot \begin{bmatrix} 0 & \sin 2\varphi - \sin\varphi & -\sin\varphi \\ \cos 2\varphi + 2\cos\varphi & \cos 2\varphi + \cos\varphi & \cos\varphi \end{bmatrix} \tag{17}$$

and the static equilibrium is defined by the following equations

$$\begin{aligned} L(\cos 2\varphi + 2\cos\varphi)\cdot F_y &= 0 \\ L(\sin 2\varphi - \sin\varphi)\cdot F_x + L(\cos 2\varphi + \cos\varphi)\cdot F_y - k\cdot 3\varphi &= 0 \\ -L\sin\varphi\cdot F_x + L\cos\varphi\cdot F_y + k\cdot 3\varphi &= 0 \end{aligned} \tag{18}$$

which do not have exact solutions corresponding to the configuration angles $(\varphi, -3\varphi, 3\varphi)$. However, for the very small $\varphi \to 0$ the above system is consistently providing the following solution

$$F_x \to 3k\varphi/L\sin\varphi, \qquad F_y = 0 \tag{19}$$

Hence, the critical force for the Z-shape case can be obtained by computing the relevant limit, which is

$$F_x^Z = \lim_{\varphi\to 0}\left(\frac{3k\varphi}{L\sin\varphi}\right) = \frac{3k}{L} \tag{20}$$

So, for the considered three-link serial manipulator, similar to the Euler column, there are two possible post-buckling manipulator shapes satisfying the static equilibrium equations. For the column, there were referred to as the half-sine and the full-sine shapes, while here they are called U- and Z-shapes respectively. Moreover, both half-sine and U-shape correspond to the stable equilibrium and relevant critical forces are lower than the alternative case of the full-sine and Z-shapes, which correspond to the unstable equilibrium. It is also worth mentioning that these conclusions are in good correspondence with the simulation results obtained via the energy approach in the previous section (see **Fig 5**). Nevertheless, despite obvious simplicity, the above-presented method can be hardly applied to the general case dealing with *n*-link serial manipulators.

## 4.2 Critical force for *n*-link manipulator

Further, let us present a general technique dealing with a *n*-link serial manipulator containing *n* elastic joints, which is shown in **Fig. 15**, where the left-hand side is fixed. It is assumed that the manipulator initial shape is straight and its geometry is described by the following equations

$$\delta_x = \sum_{j=1}^{n} L_j - \sum_{j=1}^{n}\left( L_j \cdot \cos(\sum_{i=1}^{j} q_i) \right) \qquad \delta_y = \sum_{j=1}^{n}\left( L_j \cdot \sin(\sum_{i=1}^{j} q_i) \right) \qquad (21)$$

where $L_j$ denotes the manipulator link lengths, $q_i$ is the joint rotation angles, and ($\delta x$, $\delta y$) is the deflections of the manipulator end-effector caused by the external loading $\mathbf{F} = (F_x, F_y)^T$. As known from mechanics, the manipulator equilibrium configuration must satisfy the following matrix equation

$$\mathbf{M(q)} + \mathbf{J(q)}^T \cdot \mathbf{F} = \mathbf{0} \qquad (22)$$

which includes the kinematic Jacobian matrix

$$\mathbf{J(q)} = \begin{bmatrix} -\sum_{j=1}^{n}\left( L_j \cdot \sin \sum_{i=1}^{j} q_i \right) & -\sum_{j=2}^{n}\left( L_j \cdot \sin \sum_{i=1}^{j} q_i \right) & \cdots & -\sum_{j=n}^{n}\left( L_j \cdot \sin \sum_{i=1}^{j} q_i \right) \\ \sum_{j=1}^{n}\left( L_j \cdot \cos \sum_{i=1}^{j} q_i \right) & \sum_{j=2}^{n}\left( L_j \cdot \cos \sum_{i=1}^{j} q_i \right) & \cdots & \sum_{j=n}^{n}\left( L_j \cdot \cos \sum_{i=1}^{j} q_i \right) \end{bmatrix}_{2\times n} \qquad (23)$$

and the elastic joint torque vector $\mathbf{M(q)}$, which for the straight initial configuration can be expressed as

$$\mathbf{M(q)} = -[k_1 q_1, k_2 q_2, ..., k_n q_n]^T \qquad (24)$$

where $k_1,...,k_n$ are the joint stiffness coefficients

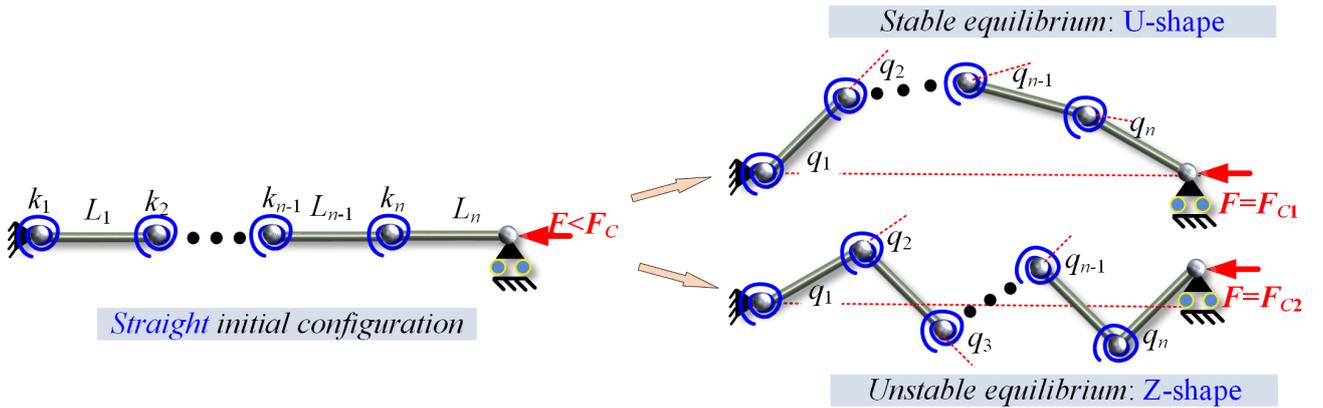

**Fig. 15** Evolution of the manipulator shape after the buckling: possible equilibriums with U-, Z- and ZU-shapes.

It should be noted that the equilibrium equation (22) provides *n* scalar relations applied to *n*+2 variables ($q_1, q_2,…, q_n$) and ($F_x, F_y$). So totally, combining (22) with two geometric equations (21), one can obtain *n*+2 nonlinear equations for *n*+2 unknowns corresponding to some given deflection ($\delta_x, \delta_y$). Obviously, in the general case, such a nonlinear system can be solved only numerically. However, for some small neighborhoods of the straight initial configuration, it is possible to find an analytical solution, which allows us to compute the critical force causing the manipulator buckling. For this computation, let us also assume that both the end-effector displacement ($\delta_x, \delta_y$) and the angles ($q_1, q_2,...,q_n$) are small enough. Under such assumption, the geometric model can be linearized and equations (21) are rewritten as

$$\delta_x = \frac{1}{2}\sum_{j=1}^{n}\left(L_j \cdot \left(\sum_{i=1}^{j}q_i\right)^2\right); \quad \delta_y = \sum_{j=1}^{n}\left(L_j \cdot \sum_{i=1}^{j}q_i\right). \tag{25}$$

Besides, in this case, the Jacobian matrix can be expressed in the form

$$\mathbf{J}(\mathbf{q}) = \begin{bmatrix} -\sum_{j=1}^{n}\left(L_j \cdot \sum_{i=1}^{j}q_i\right) & -\sum_{j=2}^{n}\left(L_j \cdot \sum_{i=1}^{j}q_i\right) & \cdots & -\sum_{j=n}^{n}\left(L_j \cdot \sum_{i=1}^{j}q_i\right) \\ \sum_{j=1}^{n}L_j & \sum_{j=2}^{n}L_j & \cdots & \sum_{j=n}^{n}L_j \end{bmatrix}_{2\times n} \tag{26}$$

that can be further presented in a more compact way

$$\mathbf{J}(\mathbf{q}) = \left[\mathbf{S}_1 \cdot \mathbf{q} \,\vdots\, \mathbf{S}_0\right]^T \tag{27}$$

with the following matrix components

$$\mathbf{S}_1 = -\begin{bmatrix} \sum_{j=1}^{n}L_j & \sum_{j=2}^{n}L_j & \cdots & \sum_{j=n}^{n}L_j \\ \sum_{j=2}^{n}L_j & \sum_{j=2}^{n}L_j & \cdots & \sum_{j=n}^{n}L_j \\ \cdots & \cdots & \cdots & \cdots \\ \sum_{j=n}^{n}L_j & \sum_{j=n}^{n}L_j & \cdots & \sum_{j=n}^{n}L_j \end{bmatrix}; \quad \mathbf{S}_0 = \begin{bmatrix} \sum_{j=1}^{n}L_j \\ \sum_{j=2}^{n}L_j \\ \cdots \\ \sum_{j=n}^{n}L_j \end{bmatrix} \tag{28}$$

The latter allows us to transform the static equilibrium equation (22) in the linearized form

$$\mathbf{K}_q \mathbf{q} + \mathbf{S}_1 \mathbf{q} \cdot F_x + \mathbf{S}_0 \cdot F_y = \mathbf{0} \tag{29}$$

that includes the vectors of unknowns $\mathbf{q}=(q_1,...,q_n)^T$, $\mathbf{F}=(F_x,F_y)^T$, the geometric parameter matrices $\mathbf{S}_0$, $\mathbf{S}_1$, and the joint stiffness matrix $\mathbf{K}_q = diag(k_1,...,k_n)$. Further, taking into account that the initial configuration is straight, let us assume that $\delta_y = 0$ and combine the above equation (29) with the corresponding geometric constraint obtained from the linearized model. It can be easily proved that such constraint $\delta_y = 0$ can be also expressed via the matrix $\mathbf{S}_0$ as

$$\mathbf{S}_0^T \mathbf{q} = 0. \tag{30}$$

This grouping leads to an extended system of $n+1$ linear equations with $n+2$ unknowns $(q_1, q_2,...,q_n)$ and $F_x$, $F_y$, which describes the loaded manipulator behavior in the neighborhood of the initial straight configuration. Using the above notations, this extended system of equations can be presented as

$$\begin{bmatrix} \mathbf{S}_1 & \mathbf{0}_{n\times 1} \\ \mathbf{0}_{1\times n} & 0 \end{bmatrix} \cdot \begin{bmatrix} \mathbf{q} \\ F_y \end{bmatrix} \cdot F_x + \begin{bmatrix} (\mathbf{K}_q)_{n\times n} & \mathbf{S}_0 \\ \mathbf{S}_0^T & 0 \end{bmatrix} \cdot \begin{bmatrix} \mathbf{q} \\ F_y \end{bmatrix} = \mathbf{0} \tag{31}$$

and rewritten further in the following matrix form

$$(\mathbf{A} \cdot F_x + \mathbf{B})\mathbf{v} = \mathbf{0} \tag{32}$$

where the notations are $\mathbf{A}$, $\mathbf{B}$ are $(n+1)\times(n+1)$ square matrices of parameters and $\mathbf{v}$ is the vector of size $(n+1)$ including unknowns, i.e.

$$\mathbf{A} = \begin{bmatrix} \mathbf{S}_1 & \mathbf{0}_{n\times 1} \\ \mathbf{0}_{1\times n} & 0 \end{bmatrix}_{(n+1)\times(n+1)}; \quad \mathbf{B} = \begin{bmatrix} (\mathbf{K}_q)_{n\times n} & \mathbf{S}_0 \\ \mathbf{S}_0^T & 0 \end{bmatrix}_{(n+1)\times(n+1)}; \quad \mathbf{v} = \begin{bmatrix} \mathbf{q} \\ F_y \end{bmatrix}_{(n+1)\times 1} \tag{33}$$

Thus, it can be easily seen that the obtained extended matrix equation (32) is similar to the basic equation considered in the classical matrix analysis for computing the matrix eigenvectors and eigenvalues. The latter allows us to find the relation between the desired critical force $F_x^0$ and the eigenvalues of the matrix $\mathbf{B}^{-1}\mathbf{A}$. In particular, taking into account that here the matrix $\mathbf{B}$ is invertible, i.e. $\det(\mathbf{B}) \neq 0$, the above equation (32) can be easily converted to the standard form used in conventional matrix algebraic

$$\left(\mathbf{B}^{-1}\mathbf{A} - \lambda \cdot \mathbf{I}\right)\mathbf{v} = \mathbf{0}; \qquad \lambda = -1/F_x \tag{34}$$

It should be noted that the above matrix equation defines only $(n+1)$ relations between $(n+2)$ unknowns because it does not include explicitly the geometric constraint (25) associated with $\delta_x$, which is assumed to be small but arbitrary. Nevertheless, besides its deficiency, the equation (34) allows us to obtain some very useful results.

Using the matrix $\mathbf{B}^{-1}\mathbf{A}$ and computing its eigenvalues/eigenvectors, one can obtain a set of possible stable and unstable equilibrium configurations corresponding to $\delta_y = 0$ and some small $\delta_x$. In such equilibrium configurations, the manipulator shape is defined by the vectors $\mathbf{q}_i$ that are extracted from the eigenvectors $\{\mathbf{v}_1,...,\mathbf{v}_{n+1}\}$, which are defined as $\mathbf{v}_i = (q_1^i, q_2^i,...,q_n^i, F_y^i)^T$. However, taking into account that the classical eigenvalue analysis assumes that the norm $\|\mathbf{v}_i\| = 1$, here the vectors $\mathbf{v}_i$ must be replaced by $\mu \cdot \mathbf{v}_i$, where $\mu$ is some small scaling factor that may be both positive and negative. It is clear that such scaling does not violate the equation (34) but allows us to satisfy the assumption that $\mathbf{q}_i$ is small enough. Using the obtained eigenvectors $\mathbf{v}_1,...,\mathbf{v}_{n-1}$, it is possible to express the joint variables $q_j$ and the components of the external force $F_x$, $F_y$ in the following way

$$\begin{aligned} q_j &= \mu \cdot v_{ij} \\ F_x &= -1/\lambda_i \qquad j=1,...,n; \quad \forall i : \lambda_i \neq 0 \\ F_y &= \mu \cdot v_{i,n+1} \end{aligned} \tag{35}$$

where $v_{ij}$ are the components of the eigenvector $\mathbf{v}_i = \left[v_{i,1},...,v_{i,n+1}\right]^T$ and $\mu \approx 0$ is an arbitrary small number. Besides, it can be proved that there are exactly two zero eigenvalues among $\lambda_i$, which correspond to a pure straight configuration. The latter follows from some specific properties of the matrix $\mathbf{B}^{-1}\mathbf{A}$, where $\det(\mathbf{A}) = 0$, $\det(\mathbf{B}) \neq 0$ and $\text{rank}(\mathbf{B}^{-1}\mathbf{A}) = n-1$

Hence, taking into account that $\mu$ can be both positive and negative, the above expressions yield $2(n-1)$ possible manipulator shapes corresponding to the static equilibriums under different axial loadings defined by the nonzero eigenvalues $\lambda_i$, $i=1,...,n-1$. It is clear that the buckling happens when the external force $F_x$ achieves the minimum of these possible equilibrium loadings, which allows us to compute the desired critical force $F_x^0$ using the largest (in absolute value) eigenvalue of the matrix $\mathbf{B}^{-1}\mathbf{A}$:

$$F_x^0 = -\frac{1}{\max|\lambda_i|} \tag{36}$$

that corresponds to the minimum energy non-straight equilibrium with non-zero configuration angles $\mathbf{q} \neq \mathbf{0}$.

The above conclusion can be also confirmed by the straightforward application of the energy method. In particular, using the expressions (32) the manipulator elastostatic energy $E = \sum_{j=1}^{n} k_j \cdot q_j^2 / 2$ in the $i$th equilibrium configuration and corresponding deflection $\delta x$ can be expressed as

$$E_i = \frac{\mu^2}{2}\sum_{j=1}^{n}\left(k_j \cdot v_{ij}^2\right) \qquad \delta_x = \frac{\mu^2}{2}\sum_{s=1}^{n} L_s \cdot \left(\sum_{j=1}^{s} v_{ij}\right)^2 \tag{37}$$

This allows us to compute the elastostatic energy corresponding to different equilibriums with the same $\delta_x$, and obtain corresponding energy-displacement relation in the neighborhood of the considered straight configuration

$$E_i = \mu_i^0 \cdot \delta_x \tag{38}$$

where $\mu_i^0$ is the coefficient which will be further referred to as the energy factor that is computed as

$$\mu_i^0 = \frac{\sum_{j=1}^{n} k_j \cdot v_{ij}^2}{\sum_{s=1}^{n} L_s \cdot \left(\sum_{j=1}^{s} v_{ij}\right)^2} \tag{39}$$

Thus, as follows from (37), for the same small deflection $\delta_x$ there are $n$-1 equilibrium shapes with different elastostatic energy $E_1,...,E_{n-1}$. It is clear that the smallest of them with minimum energy factor $\mu_i^0$ corresponds to the first buckling mode which is really observed in practice. Below the factor $\mu_i^0$ will be also used for the equilibrium stability analysis of different equilibriums, allowing us to confirm that the minimum of $\mu_i^0$ corresponds to the largest eigenvalue $\lambda_i$ (in absolute value).

The developed technique can be illustrated by an application example dealing with a trivial four-link manipulator for which the lengths of all links and the stiffness coefficients are equal to one, i.e. $n=4; L_i=1, k_i=1, \forall i$. For this manipulator, the matrices $\mathbf{S}_1$ $\mathbf{S}_0$ can be expressed as

$$\mathbf{S}_1 = -\begin{bmatrix} 4 & 3 & 2 & 1 \\ 3 & 3 & 2 & 1 \\ 2 & 2 & 2 & 1 \\ 1 & 1 & 1 & 1 \end{bmatrix}; \quad \mathbf{S}_0 = \begin{bmatrix} 4 \\ 3 \\ 2 \\ 1 \end{bmatrix} \tag{40}$$

yielding the following characteristic polynomial

$$\lambda^5 + \frac{9}{5}\lambda^4 + \frac{9}{10}\lambda^3 + \frac{2}{15}\lambda^2 = 0$$

with three non-zero roots $\lambda_i \in \{-1.0822, -0.4337, -0.2841, 0, 0\}$, whose eigenvectors are presented in **Table 2**.

**Table 2** Non-zero eigenvalues and corresponding eigenvectors of matrix $\mathbf{B}^{-1}\mathbf{A}$ for four-link manipulator

|    | $\lambda$ | $v_1$ | $v_2$ | $v_3$ | $v_4$ | $v_5$ |
|----|---------|--------|---------|---------|---------|---------|
| #1 | -1.0822 | 0.5185 | -0.0902 | -0.6156 | -0.5721 | -0.1296 |
| #2 | -0.4337 | 0.3863 | -0.6009 | -0.2026 | 0.6628 | -0.0966 |
| #3 | -0.2841 | 0.2062 | -0.5713 | 0.6623 | -0.4356 | -0.0516 |

The obtained eigenvalues and eigenvectors allow us to evaluate the manipulator shape for all possible equilibriums in the neighborhood of the initial configuration. It is clear that each $\lambda_i$ yield two symmetrical equilibrium configurations (for both $\mu<0$ and $\mu>0$), whose shape can be evaluated by analyzing the signs of $v_{ij}$. Hence, the total number of the different equilibrium shapes is equal to $2(n$-1), and two of them providing the minimum energy are globally stable and observed in practice. Examples of such shapes after buckling (obtained for very small $\delta_x$) are presented in **Table 3**. It should be mentioned that in contrast to the case when $n$=3 with U- and Z-shapes only (see subsection 4.1), there is here an additional "ZU-shape" that is also unstable. Besides, it is worth mentioning that if the compressive force is small and less than the critical value $F_x < F_x^0$, all discovered U-, Z- and ZU-shapes are unfeasible and the manipulator keeps its stable straight configuration with $\delta_x = 0$.

**Table 3** Possible manipulator shapes in static equilibrium after the buckling for the four-link manipulator (for small deflection $\delta_x \approx 0$).

|                     | q1 | q2 | q3 | q4 | Geometric shape | Stability | Energy factor $\mu_i^0$ | Axial force $Fx$ |
|---------------------|----|----|----|----|-----------------|-----------|--------------------------|------------------|
| Case #1<br>q1<0     | −  | +  | +  | +  | U shape:        | stable    | **0.9240**               | **0.9240**       |
|                     | −  | +  | −  | +  | Z shape:        | unstable  | 3.5203                   | 3.5199           |
|                     | −  | +  | +  | −  | ZU shape:       | unstable  | 2.3057                   | 2.3057           |
| Case #2<br>q1>0     | +  | −  | −  | −  | U shape:        | stable    | **0.9240**               | **0.9240**       |
|                     | +  | −  | +  | −  | Z shape:        | unstable  | 3.5203                   | 3.5199           |
|                     | +  | −  | −  | +  | ZU shape:       | unstable  | 2.3057                   | 2.3057           |

The above-proposed eigenvalues technique gives a universal method for computing the critical force causing buckling of multi-link serial manipulators. This technique is also summarised in **Algorithm 1**, which can be applied to serial manipulators with arbitrary links lengths and joint stiffness coefficients.

**Algorithm 1**. Critical force computation for multi-link serial manipulators

| | | |
|---|---|---|
| **Input:** | $\mathbf{L} = [L_1, L_2, ..., L_n]$ : | vector of manipulator links lengths |
| | $\mathbf{K}_q = [k_1, k_2, ..., k_n]$ : | vector of manipulator joints stiffness coefficients |
| **Output:** | $F_x^0$ : | value of critical force causing the buckling |

1: Compute the auxiliary matrix $\mathbf{S}_1(\mathbf{L})$ and vector $\mathbf{S}_0(\mathbf{L})$ using equation (28)
2: Obtain the principal matrices $\mathbf{A}(\mathbf{S}_1)$ and $\mathbf{B}(\mathbf{K}_q, \mathbf{S}_0)$ using equation (33)
3: Compute the eigenvalues $\boldsymbol{\lambda} = [\lambda_1, \lambda_2, ..., \lambda_{n-1}]$ of the matrix $\mathbf{B}^{-1}\mathbf{A}$
4: Estimate the critical force $F_x^0 = -1 / \max |\lambda_i|$ causing the buckling

The elastostatic energy associated with each equilibrium shape can be easily evaluated using the energy factor that is also included in **Table 3**. As follows from these data, the energy minimum corresponds to the U-shape, which has the minimum value of the energy factor $\mu_i^0$ and minimum compressive force $F_x$. These results are also confirmed by straightforward computing of the energy function $E(q_1, q_4)$ presented in **Fig. 16** that was obtained under the geometric constraints (21) allowing to exclude $q_2, q_3$. As follows from this figure, two global maximums/minimums are corresponding to the stable/unstable equilibriums respectively. Besides, two saddle points obviously correspond to unstable equilibriums. So totally, for the given deflections ($\delta_x$, $\delta_y$), there are six equilibriums: two stable ones with U-shape and four unstable ones with Z- and ZU-shapes. It is also worth stressing that in contrast to **Table 3**, the axial deflection is not very small here $\delta_x = 0.5L$, but the manipulator also prefers to keep the U-shape after the buckling with minimal elastostatic energy.

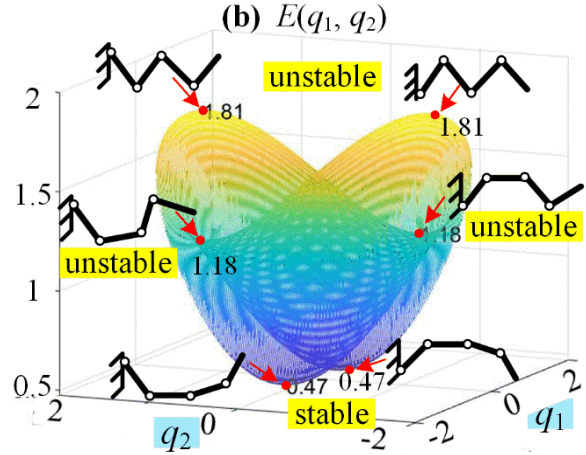

**Fig. 16** The energy function $E(q_1, q_2)$ and static equilibriums for $\delta x = 0.5L$, $\delta y = 0$.

Therefore, the proposed technique, which is based on the eigenvalue analysis of the special parameter matrix, allows easily to estimate the critical load for any *n*-links serial manipulator under the compressive loading. This method is also in agreement with the result from a subsection of 4.1 where the critical force was computed for a three-link manipulator using a particular technique.

## 5  Discussions

This work is mainly focused on the stiffness analysis of 2D planar serial chains, however, some of the obtained results can be easily adapted for 3D spatial cases. For instance, if the kinematic chain can be decomposed into several planar substructures then the critical force for the straight configuration can be straightforwardly computed by selecting minimum from corresponding critical forces for the sub-chains. Moreover, the relevant extension of static equilibrium equations and corresponding Jacobian linearization in the neighborhood of the straight configuration obviously should lead to the equation similar to (29) that is linear with respect to the joint variable vector **q**. Hence, the critical force computing for a 3D case can be also reduced to the eigenvalues and eigenvectors computing of some matrix depending on the manipulator geometric parameters. However, some difficulties may arise if the coupling between the liner and angular component influence essentially on the manipulator stiffness behavior or if non-negligible external torque is applied to the end-effector, in addition to the considered external force. Nevertheless, these issues are rather non-natural for real robotic systems.

Also, the main results were obtained for the serial manipulators with revolute joints only. Nevertheless, this limitation is not very essential from the practical point of view. In fact, since deflections in prismatic joints do not change the manipulator shape under the loading and corresponding deflections are relatively small (compared to the original link lengths), the prismatic joints do not affect the critical force value, while the total deflections should be obviously higher and the manipulator post-buckling behavior may be slightly different.

Another important issue is that in the presented work it was indirectly assumed that the links have regular symmetrical shape and their elasticity can be included in the model via the joint stiffness coefficients. However, in the case of complex shape links the corresponding $6\times6$ stiffness matrix is non-diagonal and takes into account coupling between deflections in different directions. To overcome this difficulty, VJM-based technique can be applied where the considered one-dimensional compliant joints are replaced by generalized six-dimensional virtual springs. So, it is also possible here to apply the general results presented in this paper. It is worth mentioning that the required $6\times6$ stiffness matrix for serial chain links can be estimated using virtual experiments in a CAD environment [13, 46].

It should be also mentioned that the presented results carefully analyzed both stable and unstable equilibrium of compressed kinematic chains. But in practice, unstable static equilibrium can hardly be hardly be observed since any disturbance moves the manipulator to the stable configuration. So, in order to ensure that numerical routine produces desired stable solution, it is reasonable to use the relevant stability criterion proposed in [43]. In the case when the unstable equilibrium is discovered, it is sufficient to add some small disturbance to the system configuration and find the equilibrium again. The latter is extremely important for practice.

Besides, in this study, we ignore the impact of gravity and/or gravity compensation on the manipulator stiffness behavior. In fact, this issue can be ignored if the motion plane is orthogonal to the gravity direction. Also, the gravity forces do not affect the stiffness behavior if they are directed along the axis of the straight configuration, but they definitely change critical force. However, in other cases, the gravity compensators and gravity forces affect stiffness behavior. Since the gravity compensators were not addressed in the considered mechanical structures, it is required to adjust some expression for the critical forces in the case of the gravity influence is essential. On the other hand, for the structures studied in this paper, the gravity forces do not affect the behavior essentially, their impact is comparable with small external disturbance. It is also worth mentioning that the effect of inertia forces is also out of the scope of this paper because the buckling phenomenon is usually studied for static cases only.

Finally, it is clear that in practice the model parameters are estimated with limited accuracy. This means that critical force is also estimated with some tolerance. Besides, the post-buckling behavior for real parameters (not nominal) may differ essentially from the theoretical one (another force magnitude leads to the force-deflection curve shift while other stiffness parameters change the curve shape). For this reason, it is not possible hardly rely on real applications on the post-buckling behavior predicted by the models with nominal parameters. The obtained results are mainly useful to estimate the range of the critical forces causing the buckling phenomenon. Hence, in practice, it is reasonable to limit the compressive force magnitude with some safety factor, in order to avoid undesirable manipulator behavior.

# 6  Conclusions

The paper focuses on the non-linear stiffness behavior of planar serial robotic manipulators with revolute joints under the compressive loading assuming that the manipulator's initial shape is either straight or non-straight one. It was proved that for the straight initial configuration if the compressive force exceeds some critical value, there may be observed the buckling where the manipulator suddenly changes its straight shape and stiffness properties. For computing this critical force, a general eigenvalue-based technique was proposed that can be applied to any multi-link serial manipulator. For the non-straight initial configuration, a universal energy-based technique was applied that allowed detection of quasi-buckling phenomena. It is also observed here very fast but not an instant modification of the manipulator shape and its stiffness coefficient if the compressive force exceeds some critical value. This force was computed by straightforward minimization of the manipulator strain energy for different end-effector deflections which allowed to obtain non-linear force-deflection curves with essential change of the slope in some areas. The developed technique has been successfully applied to the stiffness modeling of three- and four-link planar manipulators that are subjected to compressive forces. In the future, the developed method will be generalized for the three-dimensional case and applied to more complex architectures.

# Acknowledgment

This work was supported by the PhD China Scholarship Council (No. 201801810036) and a grant from the Russian Science Foundation (project No:19-79-10246).